\documentclass[10pt,twocolumn,letterpaper]{article}

\usepackage{cvpr}              

\newcommand{\x}{\mathbf{x}}
\newcommand{\y}{\mathbf{y}}
\newcommand{\zeroVec}{\mathbf{0}}
\newcommand{\bbD}{\mathbb{D}}

\newcommand{\cmark}{\ding{51}}
\newcommand{\xmark}{\ding{55}}

\usepackage{overpic}
\usepackage{pifont}
\usepackage[accsupp]{axessibility}

\definecolor{cvprblue}{rgb}{0.21,0.49,0.74}
\usepackage[pagebackref,breaklinks,colorlinks,allcolors=cvprblue]{hyperref}

\def\paperID{15} 
\def\confName{CVPR}
\def\confYear{2026}

\title{Semantic Alignment in Hyperbolic Space for \\  Open-Vocabulary Semantic Segmentation}

\renewcommand{\and}{\hspace{2em}}

\author{
Hoang M. Truong \and
Hai Nguyen-Truong \and
Dang Huynh\thanks{Corresponding author: \texttt{dang.huynh@fulbright.edu.vn}} \\
Fulbright University Vietnam  \\ 
{\small{\url{https://tmhoanggg.github.io/HyRo/}}}
}

\begin{document}
\maketitle
\begin{abstract}
Open-vocabulary semantic segmentation requires adapting image-level vision-language models such as CLIP to dense pixel-level prediction, which is challenging due to the mismatch between hierarchical structure and semantic alignment in the embedding space. While recent works leverage hyperbolic geometry to model hierarchical relationships, they align embeddings across hierarchical levels but overlook semantic misalignment among embeddings within the same level. In this work, we propose HyRo, a hyperbolic fine-tuning framework that decouples hierarchical and semantic alignment in the Poincar\'e ball model. HyRo aligns hierarchical levels by adjusting the hyperbolic radius and refines semantic relationships through angular alignment using an orthogonal transformation that theoretically preserves the hyperbolic radius. Experiments on standard open-vocabulary semantic segmentation benchmarks demonstrate that HyRo achieves state-of-the-art performance over prior methods.
\end{abstract}    
\section{Introduction}
\label{sec:intro}
Open-vocabulary semantic segmentation aims to assign each pixel in an image to a semantic category specified by textual descriptions, including categories unseen during training. To tackle this challenging task, vision-language foundation models such as CLIP~\cite{CLIP} have emerged as powerful tools, as they are trained on large-scale image–text pairs and capture rich semantic alignments between visual and linguistic modalities. Consequently, CLIP is widely adopted for open-vocabulary learning by representing class names as text embeddings, effectively enabling language-driven classification for downstream tasks. However, CLIP is originally pre-trained for image-level representation learning, and thus requires additional adaptation to produce dense, pixel-level predictions for semantic segmentation.

To achieve this adaptation, early approaches~\cite{odise, mask-adapted-clip, zhou2022maskclip, SAN, xu2022simple} decoupled the problem into two stages, utilizing mask proposal generators followed by a pre-trained CLIP model to classify the resulting regions. While such methods benefit from explicit object localization and clear boundaries, the reliance on pre-defined mask proposals introduces a closed-set bias that can limit scalability and generalization to unseen categories. Recent methods~\cite{catseg, dpseg, SED} instead directly fine-tune CLIP within a shared representation space, enabling dense pixel-level predictions while preserving the semantic richness of the original CLIP representation.

\begin{figure}[t]
    \centering
    \includegraphics[width=1\linewidth]{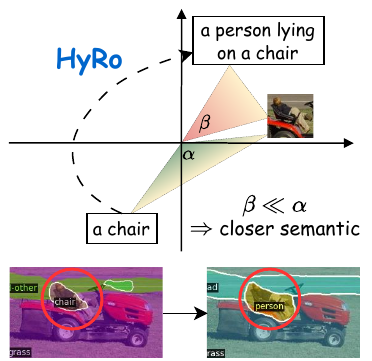}\vspace{-5pt}
    \caption{HyRo rotates the text embeddings to achieve a smaller angle ($\beta$) relative to the target image embeddings compared to the initial angle ($\alpha$). This geometric adjustment enables the model to resolve semantic ambiguities and produce more accurate, fine-grained segmentations.}
    \label{fig:intro_hyro}
    \vspace{-8pt}
\end{figure}

Notably, hyperbolic space has recently gained attention for its ability to capture intrinsic hierarchical structures. Several works~\cite{meru, hycoclip} incorporate hierarchical representations into CLIP-like vision–language models by embedding visual and textual entities in hyperbolic space, enabling the model to better capture hierarchical relationships between modalities. In contrast, a recent advancement in open-vocabulary semantic segmentation, HyperCLIP~\cite{hyperclip}, observes that the hierarchy of image embeddings shifts from an image-level to a pixel-level during fine-tuning. To explain this, they integrate the Poincar\'e ball model to adjust the hyperbolic radius of text embeddings to match this pixel-level granularity. However, a critical limitation of this approach is the absence of explicit constraints on the semantic alignment between embeddings, which is essential for effective open-vocabulary learning. Semantic relationships can be represented in hyperbolic space through geometric structures such as imaginary cones, as explored in prior work~\cite{meru, hycoclip, hyperbolic_iccv_oral}, or equivalently through angular relationships at the origin.

To address this key limitation, we introduce HyRo (Hyperbolic Rotation), a novel fine-tuning strategy that explicitly disentangles hierarchical and semantic information in hyperbolic space. While prior work primarily focuses on the \textit{hierarchical geometry} of the embedding space, our approach explicitly optimizes the \textit{semantic} aspect. HyRo is motivated by the observation that hierarchy and semantics correspond to distinct geometric properties, as hierarchy is encoded by radial distance, while semantic similarity is reflected in angular orientation.~\Cref{fig:intro_hyro} illustrates a failure case in prior work and demonstrates how HyRo resolves it. Specifically, prior work~\cite{hyperclip} fine-tunes solely via radius scaling and overlooks semantic orientation. This potentially leads to semantically dissimilar concepts being embedded at similar radii but incorrect angles, degrading the discrimination between categories. For instance, in an image depicting a person lying on a chair, HyperCLIP loses semantic understanding and misclassifies both entities as a single ``chair" object. HyRo addresses this by rotating the initial embedding to achieve improved angular alignment with related embeddings. After this rotation, the model better segments the scene, correctly recognizing the person as the distinct foreground subject.

We theoretically demonstrate that orthogonal transformations in the Poincar\'e ball model act as ideal rotation operations (see~\cref{subsec:angle_adjustment}). These transformations enable angular adjustment of embeddings, refining semantic alignment without altering their radius, thereby preserving the hierarchical structure established by prior radius tuning. By decoupling these geometric properties, HyRo enables CLIP to simultaneously maintain the pixel-level granularity required for segmentation and enhance the semantic discrimination necessary for open-vocabulary generalization. Overall, our approach highlights the importance of explicitly modeling both hierarchical and angular relationships when adapting vision–language representations for dense open-vocabulary prediction.

Our contributions are summarized as follows:
\begin{itemize}
    \item We propose HyRo, an orthogonal transformation strategy that adjusts angular relationships in hyperbolic space while preserving hierarchical structure, with theoretical justification provided.
    \item We introduce a hyperbolic fine-tuning framework that decouples hierarchical alignment (radius) and semantic refinement (angle) for open-vocabulary semantic segmentation.
    \item We demonstrate that HyRo achieves state-of-the-art performance on standard benchmarks, validating the effectiveness of our geometric approach.
\end{itemize}
\section{Related Works}
\label{sec:related_works}

\subsection{Open-Vocabulary Semantic Segmentation.}
Previous works~\cite{SED, catseg, SAN, dpseg, zhou2022maskclip} on open-vocabulary semantic segmentation aim to adapt foundation vision–language models pre-trained on large-scale datasets, such as CLIP~\cite{CLIP}, for pixel-level prediction. This adaptation requires transforming high-level vision–language representations into dense predictions, which poses a key challenge for the task. Early approaches directly fine-tune the encoders of CLIP; however, prior methods~\cite{SAN, yu2023fcclip, zhou2022maskclip} observe overfitting to seen classes, as such fine-tuning can degrade the generalization ability of CLIP. As a result, many works~\cite{zegformer-decoupling-zero-shot-ss, ghiasi2022scaling, xu2022simple, mask-adapted-clip} freeze the CLIP encoders to preserve their generalization.

More recent studies~\cite{catseg, dpseg, SED} show that fine-tuning CLIP within cost aggregation–based frameworks can alleviate this issue. Beyond Euclidean representations, several works~\cite{hyperclip, hyperspherical_ovss} further explore alternative geometric spaces for open-vocabulary semantic segmentation, providing promising insights into modeling complex semantic structures across vision and language. While existing hyperbolic methods primarily optimize the radius of embeddings to capture hierarchical relationships, the angle governing semantic similarity remains largely unexplored. We address this gap by explicitly modeling angular relationships while preserving the hierarchical structure.

\subsection{Hyperbolic Deep Learning.} 
Hyperbolic geometry provides a natural framework for representing hierarchical structures due to its exponential volume growth, which mirrors the branching nature of tree-like hierarchies~\cite{gromov1987}. In contrast, Euclidean space exhibits polynomial volume growth, making it less suitable for embedding hierarchical data with low distortion~\cite{Matousek99}. This property has led to widespread adoption of hyperbolic spaces in natural language processing~\cite{hyperbolic_nlp_1, hyperbolic_nlp_2, hyperbolic_nlp_3, hyperbolic_nlp_4}, where concepts are naturally organized in taxonomies and can be embedded as tree graphs with minimal distortion~\cite{distortion_1, distortion_2}.

Visual data also exhibits inherent hierarchical structures~\cite{vision_hierarchy}, spanning multiple levels from pixels and patches to objects and scenes. This observation has motivated recent works~\cite{hyperclip, hyperbolic_image_embeddings, hyperbolic_vision_transformers} to leverage hyperbolic geometry for visual understanding. In the context of vision-language learning, hyperbolic spaces offer a unified framework to model the hierarchical relationships between visual and textual modalities. Recent approaches~\cite{meru, hycoclip} hypothesize that textual concepts are generally more abstract than visual features and employ hyperbolic embeddings to capture this semantic hierarchy. In this work, we adopt the Poincar\'e ball model to bridge the hierarchical gap and semantic relationships between vision and language embeddings, enabling more effective alignment for open-vocabulary semantic segmentation.

\section{Methodology}
\label{sec:method}
We introduce HyRo, a method for open-vocabulary semantic segmentation that improves semantic alignment. HyRo refines semantic relationships through controlled angular transformations in hyperbolic space while preserving hierarchical information encoded in feature radii.
Specifically, we first review essential background on the Poincar\'e ball model in~\cref{subsec:background}, then present our proposed hyperbolic rotation module in~\cref{subsec:angle_adjustment}. Finally, we describe the overall architecture in~\cref{subsec:architecture}, and our decoder in~\cref{subsec:decoder}.

\subsection{Background: The Poincar\'e Ball Model}
\label{subsec:background}
We briefly review the hyperbolic geometry concepts required for our method, focusing on the Poincar\'e ball model.
The $m$-dimensional Poincar\'e ball with curvature $-c$ ($c>0$) is defined as $\mathbb{D}_c^m := \left\{ \mathbf{x} \in \mathbb{R}^m \mid c\|\mathbf{x}\|^2 < 1 \right\}$, which has radius $1/\sqrt{c}$.
Its exponential volume growth makes it naturally suited for representing hierarchical relationships~\cite{hyperbolic_nlp_1, hyperbolic_nn}.

\paragraph{Angle at the origin.} 
A key property of the Poincar\'e ball model is its conformality: angles measured at the origin are preserved from Euclidean space. Therefore, the angle $\alpha$ between two points $\mathbf{x}, \mathbf{y} \in \mathbb{D}^n_c$ at the origin is given by:
\begin{equation}
    \cos(\alpha) = \frac{\langle \mathbf{x}, \mathbf{y} \rangle}{\|\mathbf{x}\| \|\mathbf{y}\|},
    \label{eq:angle}
\end{equation}
where $\langle \cdot, \cdot \rangle$ denotes the Euclidean inner product.

\begin{figure}
    \centering
    \includegraphics[width=1\linewidth]{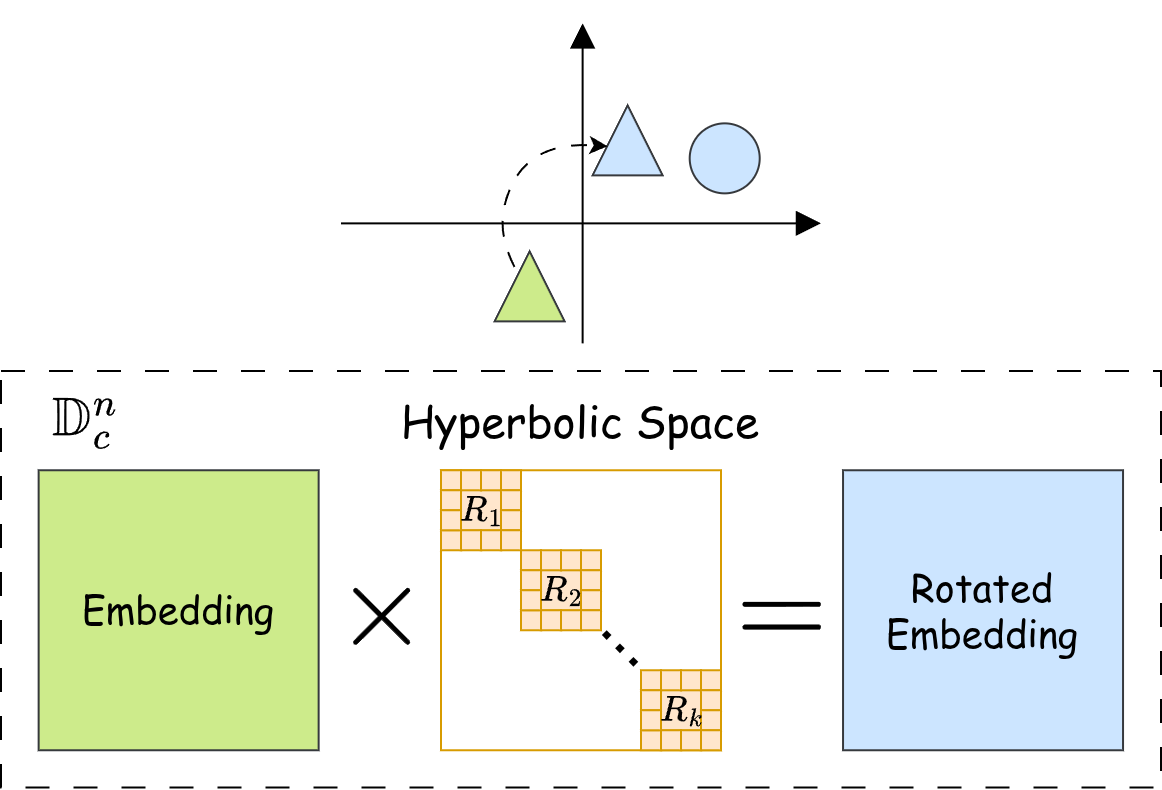}
    \caption{Overview of HyRo. Embeddings are rotated around the origin in hyperbolic space using an orthogonal block matrix to minimize the angle between visual and textual features while preserving their hyperbolic radii, thereby enhancing cross-modal semantic alignment.}
    \label{fig:hyro}
\end{figure}

\paragraph{Möbius matrix-vector multiplication.}
For $\mathbf{M} \in \mathbb{R}^{m \times m}$ and $\mathbf{x} \in \mathbb{D}_c^m$ such that $\mathbf{M} \mathbf{x} \neq \mathbf{0}$, denoting $\widetilde{\mathbf{x}} = \mathbf{M}\mathbf{x}$, the Möbius matrix-vector multiplication is defined as:
\begin{equation}
    \mathbf{M} \otimes_c \mathbf{x} = \frac{1}{\sqrt{c}} \tanh\left(
        \frac{\|\widetilde{\mathbf{x}}\|}{\|\mathbf{x}\|}\tanh^{-1}\left(\sqrt{c}\|\mathbf{x}\|\right)
    \right)\frac{\widetilde{\mathbf{x}}}{\|\widetilde{\mathbf{x}}\|}.
    \label{eq:mobius_multiplication}
\end{equation}

\paragraph{Exponential and logarithmic maps.}
Hyperbolic space is a curved manifold where standard Euclidean operations (e.g., addition, linear transformations) are not directly applicable. To bridge this gap, we use exponential and logarithmic maps that convert between Euclidean tangent space and the hyperbolic manifold.

The exponential map $\exp_\mathbf{x}^{\mathbb{D}, c}$ takes a Euclidean vector $\mathbf{v}$ from the tangent space at a point $\mathbf{x}\in\mathbb{D}_c^m$ and projects it onto the Poincar\'e ball along a geodesic (the hyperbolic equivalent of a straight line). Conversely, the logarithmic map $\log_\mathbf{x}^{\mathbb{D}, c}$ maps a point $\mathbf{y}\in\mathbb{D}_c^m$ on the manifold back to the tangent space at $\mathbf{x}$. At the origin $\mathbf{0}$, these maps have particularly simple forms:
\begin{equation}
    \exp_\mathbf{0}^{\mathbb{D}, c} (\mathbf{v}) = \frac{1}{\sqrt{c}}\tanh\left(\sqrt{c} \|\mathbf{v}\|\right)\frac{\mathbf{v}}{\|\mathbf{v}\|},
    \label{eq:exp_mapping}
\end{equation}
\begin{equation}
    \log_\mathbf{0}^{\mathbb{D}, c} (\mathbf{y}) = \frac{1}{\sqrt{c}}\tanh^{-1}\left(\sqrt{c} \|\mathbf{y}\|\right) \frac{\mathbf{y}}{\|\mathbf{y}\|}.
    \label{eq:log_mapping}
\end{equation}
These maps act as nonlinear scaling operations that preserve direction while mapping between spaces. This enables us to perform operations in the Euclidean tangent space and then map the results back to the hyperbolic manifold.

\begin{figure*}[t]
    \centering
    \includegraphics[width=1\linewidth]{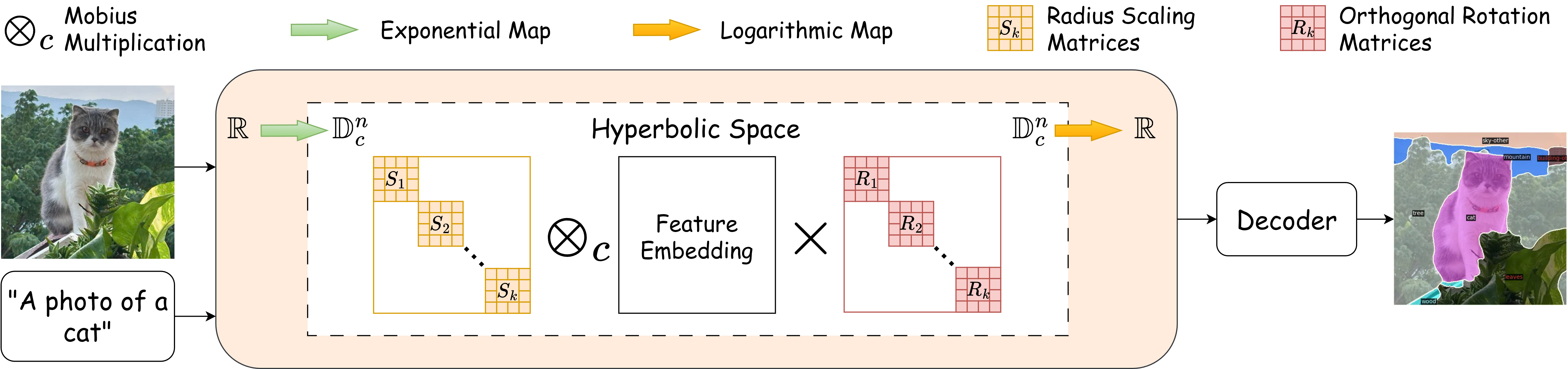}
    \caption{\textbf{Overall architecture of HyRo.} Given image and text inputs, Euclidean embeddings are mapped to the Poincar\'e ball via the exponential map. HyRo then decouples alignment into two stages: (1) \textit{Hierarchical Adjustment} using block-diagonal radius scaling matrices to align granularity, and (2) \textit{Semantic Refinement} using orthogonal rotation matrices to adjust angular relationships without altering the radius. The refined hyperbolic embeddings are mapped back to the tangent space for decoding.}
    \label{fig:architecture}
\end{figure*}

\subsection{Semantic Refinement via Rotation Matrix}
\label{subsec:angle_adjustment}
Intuitively, rotating embeddings around the origin modifies their angular relationships (semantic similarity) while preserving their radial distance (hierarchical level).
We therefore introduce an orthogonal rotation matrix to refine the angular alignment between feature embeddings while preserving their hyperbolic radii, thereby improving semantic alignment across modalities. This transformation is depicted in~\cref{fig:hyro}. Specifically, given a hyperbolic embedding $\mathbf{q} \in \bbD^d_c$ and an orthogonal matrix $\mathbf{R}$, the resulting embedding $\mathbf{v} \in \bbD_c^d$ after adjusting the angle is 
\begin{equation}
    \mathbf{v} = \mathbf{R} \mathbf{q}.
    \label{eq:v_angle_refined}
\end{equation}
Since the Poincar\'e ball model is conformal, angles measured at the origin coincide with their Euclidean counterparts.

To ensure strict orthogonality, we employ the Cayley transform~\cite{cayley_transform} to parameterize a learnable unconstrained matrix $\mathbf{\Theta} \in \mathbb{R}^{n \times n}$:
\begin{align}
    \mathbf{A} &= \mathbf{\Theta} - \mathbf{\Theta}^\top, \\
    \mathbf{R} &= \left(\mathbf{I} + \mathbf{A}\right)\left(\mathbf{I} - \mathbf{A}\right)^{-1},
\end{align}
where $\mathbf{A}$ is a skew-symmetric matrix and $\mathbf{I}$ denotes the identity matrix. This parameterization guarantees that $\mathbf{R}$ satisfies the orthogonality constraint $\mathbf{R}^\top \mathbf{R} = \mathbf{I}$. 

For computational efficiency, we adopt a block-diagonal structure inspired by~\cite{hyperspherical_ovss}. A naive application of the Cayley transform to a full $d \times d$ matrix incurs an $\mathcal{O}(d^3)$ matrix inversion cost, which becomes prohibitive for high-dimensional CLIP embeddings. Furthermore, CLIP encoders produce embeddings of different dimensionalities across modalities (\eg, 768 for vision and 512 for text in ViT-B/16), making a single shared full matrix infeasible. By decomposing $\mathbf{R}$ into $K_\mathbf{R} = d/n$ independent blocks:
\begin{equation}
    \mathbf{R} = \text{diag}\left(\mathbf{R}_1, \mathbf{R}_2, \ldots, \mathbf{R}_{K_\mathbf{R}}\right),
\end{equation}
where each block $\mathbf{R}_i \in \mathbb{R}^{n \times n}$ is constructed via the Cayley transform, the inversion cost reduces from $\mathcal{O}(d^3)$ to $K_\mathbf{R} \cdot \mathcal{O}(n^3) = \mathcal{O}(d^3/n^2)$, while also enabling parallelization across blocks. Note that setting $n = d$ (\ie, $K_\mathbf{R} = 1$) recovers a full unconstrained rotation matrix, while smaller $n$ imposes stronger structural constraints with fewer learnable parameters. We ablate this choice in~\cref{subsec:ablation}.

\paragraph{Theoretical Justification.}
We formally justify that applying an orthogonal transformation to a point in the Poincar\'e ball induces a rotation of the angle at the origin without altering the hyperbolic radius.

Let $\x \in \mathbb{D}_c^d$. Its representation in the tangent space at the origin is $\mathbf{v}_\x = \log_\zeroVec^c(\x)$. Applying an orthogonal matrix $\mathbf{R} \in \mathrm{O}(d)$ to this tangent vector yields $\mathbf{v}_\x' = \mathbf{R} \mathbf{v}_\x$. Since $\mathbf{R}$ preserves the Euclidean norm, we have $\|\mathbf{v}_\x'\| = \|\mathbf{v}_\x\|$.

Mapping $\mathbf{v}_\x'$ back to the manifold via the exponential map:
\begin{align}
    \x' &= \exp_\zeroVec^c(\mathbf{v}_\x') \nonumber \\
    &= \frac{1}{\sqrt{c}} \tanh\left(\sqrt{c}\|\mathbf{v}_\x'\|\right) \frac{\mathbf{v}_\x'}{\|\mathbf{v}_\x'\|} \nonumber \\
    &= \mathbf{R} \left[ \frac{1}{\sqrt{c}} \tanh\left(\sqrt{c}\|\mathbf{v}_\x\|\right) \frac{\mathbf{v}_\x}{\|\mathbf{v}_\x\|} \right] \nonumber \\
    &= \mathbf{R} \exp_\zeroVec^c(\mathbf{v}_\x) \nonumber \\
    &= \mathbf{R}\x.
    \label{eq:simple_rotation}
\end{align}
This confirms that the transformation simplifies to a direct matrix multiplication on the coordinates. Furthermore, the angle $\alpha'$ between the rotated point $\x'$ and a target $\y$ becomes:
\begin{equation}
    \cos(\alpha') = \frac{\langle \x', \y \rangle}{\|\x'\| \|\y\|} 
    = \frac{\langle \mathbf{R}\x, \y \rangle}{\|\x\| \|\y\|}.
    \label{eq:angle_adjusted}
\end{equation}
Crucially, since $\|\x'\| = \|\mathbf{R}\x\| = \|\x\|$ (due to orthogonality), the hyperbolic radius $\text{Rad}_{\x'} = \text{Rad}_{\x}$ remains unchanged. Thus, HyRo enables independent control over semantic alignment (angle) and hierarchical depth (radius).


\subsection{Overall Architecture}
\label{subsec:architecture}
\Cref{fig:architecture} illustrates our overall architecture.
We first map the input Euclidean feature embedding $\mathbf{z} \in \mathbb{R}^{d}$ into the Poincar\'e ball model. We utilize the origin ($\mathbf{0}$) as the reference point, as it represents the root of the hierarchy~\cite{hyperbolic_nn}. The mapping is performed via the exponential map:
\begin{equation}
    \mathbf{h} = \exp_\mathbf{0}^{\mathbb{D},c}(\mathbf{z}),
\end{equation}
where $d$ is the feature dimension, and $\mathbf{h} \in \mathbb{D}_c^d$ is the resulting hyperbolic embedding.

Once projected, we adjust the hyperbolic radius of $\mathbf{h}$ using a learnable diagonal matrix $\mathbf{S}$ introduced in~\cite{hyperclip}. In hyperbolic space, linear transformations are realized via M\"obius matrix-vector multiplication. The adjusted embedding $\mathbf{q} \in \mathbb{D}_c^d$ is obtained as:
\begin{equation}
    \mathbf{q} = \mathbf{S} \otimes_c \mathbf{h}.
    \label{eq:q_radius_adjusted}
\end{equation}

To balance expressiveness and computational efficiency, we impose a block-diagonal structure on $\mathbf{S}$. Specifically, $\mathbf{S}$ is composed of $K_\mathbf{S}$ independent sub-matrices:
\begin{equation}
    \mathbf{S} = \text{diag}(\mathbf{S}_1, \mathbf{S}_2, \dots, \mathbf{S}_{K_\mathbf{S}}),
    \label{eq:radius_diagonal_matrix}
\end{equation}
where each block $\mathbf{S}_k \in \mathbb{R}^{b \times b}$ is a learnable parameter, and the number of blocks is given by $K_\mathbf{S}=d/b$. This structure allows the model to learn distinct scaling factors for different feature subspaces.

Then, given the radius-adjusted embedding $\mathbf{q} \in \bbD^d_c$ obtained from~\cref{eq:q_radius_adjusted}, we apply an orthogonal matrix transformation as in~\cref{eq:v_angle_refined} to obtain the semantically refined embedding $\mathbf{v} \in \bbD_c^d$.

After refinement, the embeddings are mapped back to the Euclidean tangent space using the logarithmic map to facilitate compatibility with the decoder:
\begin{equation}
    \mathbf{e} = \log_{\mathbf{0}}^{\mathbb{D},c}(\mathbf{v}).
\end{equation}

In general, given a Euclidean feature embedding $\mathbf{x} \in \mathbb{R}^d$ from the visual encoder, we apply semantic refinement in hyperbolic space through the following procedure:
\begin{equation}
    \mathbf{x}' = \log_\mathbf{0}^{\mathbb{D}, c}\left(\mathbf{R} \cdot \left( \mathbf{S} \otimes_c  \exp_\mathbf{0}^{\mathbb{D},c}(\mathbf{x})\right)\right),
    \label{eq:hyp_rotation}
\end{equation}
where $\exp_\mathbf{0}^{\mathbb{D},c}$ maps the feature to the Poincaré ball, the scaling matrix $\mathbf{S}$ adjusts the hyperbolic radius, the rotation $\mathbf{R}$ refines angular relationships while preserving hyperbolic radii (and thus hierarchical information), and $\log_\mathbf{0}^{\mathbb{D}, c}$ projects back to Euclidean space for subsequent processing.

After getting the refined embeddings, we use the decoder (see~\cref{subsec:decoder}) to generate dense pixel-level predictions.

\subsection{Cost Aggregation Decoder}
\label{subsec:decoder}
We adopt the cost aggregation decoder introduced by CAT-Seg~\cite{catseg} to adapt CLIP~\cite{CLIP} for dense open-vocabulary segmentation. Instead of directly predicting pixel labels, the decoder aggregates similarity scores between visual and textual embeddings to produce pixel-level predictions.

Dense visual and textual embeddings are first extracted using the CLIP encoders~\cite{CLIP}. These embeddings are refined through the hyperbolic refinement stage described in~\cref{subsec:architecture}. The resulting refined embeddings are denoted as $D^V \in \mathbb{R}^{(H \times W) \times d}$ and $D^L \in \mathbb{R}^{N_{\mathcal{C}} \times d}$, where $H \times W$ is the spatial resolution and $N_{\mathcal{C}}$ is the number of candidate classes. Then, a cost volume $C \in \mathbb{R}^{(H \times W) \times N_{\mathcal{C}}}$ is constructed by computing cosine similarity between pixel and text embeddings:
\begin{equation}
C(i,n)=\frac{D^{V}(i)\cdot D^{L}(n)}{\|D^{V}(i)\|\,\|D^{L}(n)\|}.
\end{equation}

The cost volume is then projected into a higher-dimensional embedding space using a convolution layer, producing $F \in \mathbb{R}^{(H \times W) \times N_{\mathcal{C}} \times d_F}$.

To exploit both spatial and semantic relationships, cost aggregation is decomposed into spatial and class aggregation modules, which enforce spatial consistency and suppress background noise. For each class $n$, spatial aggregation refines the cost map using Swin Transformer blocks~\cite{swin_transformer} with window and shifted-window attention:
\begin{equation}
F^{\prime}(:,n)=\mathcal{T}^{sa}(F(:,n)).
\end{equation}

Class aggregation models relationships among category tokens. A transformer layer without positional encoding is applied across class tokens:
\begin{equation}
F^{\prime\prime}(i,:)=\mathcal{T}^{ca}(F^{\prime}(i,:)).
\end{equation}
A linear transformer is used for efficiency when handling a large number of classes.

To further improve aggregation, the original visual and textual embeddings ($D^V$ and $D^L$) are used as guidance. Projected embeddings are concatenated with cost features during attention:
\begin{equation}
F^{\prime}(:,n)=\mathcal{T}^{sa}([F(:,n);\mathcal{P}^{V}(D^{V})])
\end{equation}
\begin{equation}
F^{\prime\prime}(i,:)=\mathcal{T}^{ca}([F^{\prime}(i,:);\mathcal{P}^{L}(D^{L})]).
\end{equation}

Finally, a lightweight upsampling decoder produces high-resolution predictions. The aggregated cost volume is progressively upsampled and fused with intermediate features from the CLIP image encoder (e.g., layers 4 and 8 in ViT-B/16). Each stage performs bilinear upsampling, concatenation with upsampled CLIP features, and a $3\times3$ convolution. Starting from $24\times24$ features, the decoder progressively upsamples to $96\times96$ before the prediction head outputs the final segmentation map.

\begin{table*}[ht]
    \begin{center}
        \small
        \tabcolsep=0.08cm   
        \resizebox{\textwidth}{!}{
        \begin{tabular}{l|ccc|ccccc|c}
            \toprule

            Model & VLM & Additional Backbone & Fine-tuning Space & \texttt{A-847} & \texttt{PC-459} & \texttt{A-150} & \texttt{PC-59} & \texttt{PAS-20} & \texttt{PAS-20$^b$} \\
            \midrule
            ZS3Net~\cite{z3c}  & - & ResNet-101  & E & - & - & - & 19.4 & 38.3 & - \\
            LSeg~\cite{Lseg_ICLR_2022} & CLIP ViT-B/32 & ResNet-101   & E & - & - & - & - & 47.4 & - \\
            ZegFormer~\cite{zegformer-decoupling-zero-shot-ss} & CLIP ViT-B/16 & ResNet-101   & E & 4.9 & 9.1 & 16.9 & 42.8 & 86.2 & 62.7 \\
            ZSseg~\cite{ZSseg_ECCV_2022} & CLIP ViT-B/16 & ResNet-101  & E & 7.0 & - & 20.5 & 47.7 & 88.4 & - \\
            OpenSeg~\cite{Openseg_ECCV_2022} & ALIGN & ResNet-101  & E & 4.4 & 7.9 & 17.5 & 40.1 & - & 63.8 \\
            OVSeg~\cite{OVSeg_CVPR_2023} & CLIP ViT-B/16 & ResNet-101c  & E & 7.1 & 11.0 & 24.8 & 53.3 & 92.6 & - \\
            ZegCLIP~\cite{segclip_CVPR_2023} & CLIP ViT-B/16 & -  & E & - & - & - & 41.2 & 93.6 & - \\
            SED~\cite{SED} & CLIP ConvNeXt-B & - & E & {11.4} & \underline{18.6} & \underline{31.6} & \textbf{57.3} & {94.4} & - \\


            SAN~\cite{SAN} & CLIP ViT-B/16 & Side Adapter   & E & {10.1} & 12.6 & 27.5 & 53.8 & 94.0 & - \\
            HyperCLIP$^*$~\cite{hyperclip} & CLIP ViT-B/16 & - & {H} & \underline{11.9} & {18.2} & \textbf{31.7} & \underline{57.1} & {94.9} & \textbf{77.1} \\
            HyRo (Ours) & CLIP ViT-B/16 & - & {H} & \textbf{12.0} & \textbf{18.9} & {31.2} & \textbf{57.3} & \textbf{95.0} & \underline{76.7} \\
            \bottomrule
        \end{tabular}
        }
        \caption{\textbf{Comparison with state-of-the-art methods on standard benchmarks.} The best-performing results are presented in bold, while the second-best results are underlined. ``E'': Euclidean Space. ``H'': Hyperbolic Space.}
        \vspace{-10pt}
        \label{tab:main_table}
    \end{center}
\end{table*}

\paragraph{Training Objectives.} Following standard practice in open-vocabulary semantic segmentation methods~\cite{SED, hyperclip, SAN}, we train our model using pixel-wise cross-entropy loss. Given the predicted segmentation logits $\hat{Y} \in \mathbb{R}^{H \times W \times N_{\mathcal{C}}}$ and ground truth labels $Y \in \{1, \ldots, N_{\mathcal{C}}\}^{H \times W}$, the loss is defined as:
\begin{equation}
\mathcal{L} = -\frac{1}{H \times W} \sum_{i=1}^{H \times W} \log \frac{\exp(\hat{Y}_{i,y_i})}{\sum_{n=1}^{N_{\mathcal{C}}} \exp(\hat{Y}_{i,n})},
\end{equation}
where $y_i$ denotes the ground truth class label at spatial position $i$.

During training, we only fine-tune the hyperbolic transformation parameters (radius scaling matrices $\mathbf{S}$ and rotation matrices $\mathbf{R}$), while keeping the CLIP encoders frozen to preserve their generalization ability.
\section{Experiments}
\label{sec:experiments}

\subsection{Experimental Setup}
\paragraph{Dataset.} We evaluate our approach under the standard open-vocabulary semantic segmentation protocol following prior work~\cite{catseg}. The model is trained on COCO-Stuff~\cite{coco-stuff} and evaluated on multiple benchmark datasets, including ADE20K~\cite{ade20k}, PASCAL VOC~\cite{pascal-voc}, and PASCAL-Context~\cite{mottaghi2014role}.
ADE20K contains 20K training images and 2K validation images and is evaluated using two label sets: \texttt{A-150}, which contains the 150 most frequent classes, and \texttt{A-847}, which covers all 847 categories~\cite{zegformer-decoupling-zero-shot-ss}. PASCAL-Context includes 5K images for training and validation, with results reported on both the full 459-class setting (\texttt{PC-459}) and the 59 most frequent classes (\texttt{PC-59}). PASCAL VOC contains 20 foreground object classes and one background class; we report results on \texttt{PAS-20} following standard practice. Additionally, \texttt{PAS-20$^b$} is reported, where background labels are defined as categories present in \texttt{PC-59} but absent from \texttt{PAS-20}, following~\cite{ghiasi2022scaling}.
\vspace{-4pt}

\paragraph{Metrics.} We use mean Intersection-over-Union (mIoU) for evaluation, consistent with prior open-vocabulary semantic segmentation works~\cite{SED, SAN, OVSeg_CVPR_2023, Openseg_ECCV_2022}.
\vspace{-4pt}

\paragraph{Implementation Details.}
We fine-tune the CLIP~\cite{CLIP} ViT-B/16 model following the training protocol of~\cite{hyperclip}. AdamW~\cite{adamw} is used as the optimizer, with a learning rate of $2\times10^{-4}$ for our hyperbolic transformation parameters and $1\times10^{-6}$ for the CLIP encoder. We set the block size to $256$ for both the diagonal radius-scaling matrix and the orthogonal rotation matrix. Following common practice in open-vocabulary segmentation~\cite{catseg, hyperclip}, we use a small batch size of 8 (distributed across 8 NVIDIA A100 GPUs, 1 sample per GPU) to preserve CLIP's generalization capability. Training is performed for 40,000 iterations, taking approximately 8 hours.

\subsection{Main Results}
\paragraph{Quantitative Results.}
\Cref{tab:main_table} compares our method with state-of-the-art open-vocabulary semantic segmentation approaches across multiple benchmarks. HyRo achieves the best performance on four out of six benchmarks, demonstrating the effectiveness of hyperbolic rotation for improving semantic alignment.

The improvements on large-vocabulary benchmarks such as \texttt{A-847} and \texttt{PC-459} highlight HyRo’s ability to capture fine-grained semantic relationships among many visually similar categories. In particular, HyRo achieves 12.0 mIoU on \texttt{A-847} and 18.9 mIoU on \texttt{PC-459}. On commonly reported settings, our method obtains 31.2 mIoU on \texttt{A-150}, 57.3 mIoU on \texttt{PC-59}, 95.0 mIoU on \texttt{PAS-20}, and 76.7 mIoU on \texttt{PAS-20$^b$} matching or surpassing strong baselines~\cite{SED, hyperclip}.

Compared with the prior hyperbolic method HyperCLIP~\cite{hyperclip}, our approach improves performance on several benchmarks, including \texttt{A-847} (+0.1 mIoU), \texttt{PC-459} (+0.7 mIoU), \texttt{PC-59} (+0.2 mIoU), and \texttt{PAS-20} (+0.1 mIoU). These improvements validate that explicitly learning angular transformations, rather than relying solely on hyperbolic radius scaling, better capture the semantic relationships and the complex hierarchical structure of visual concepts. Overall, these results indicate that refining angular relationships through hyperbolic rotations improves semantic alignment and enables more effective adaptation of vision–language representations for dense open-vocabulary prediction.

\begin{figure*}[t]
  \centering
  \begin{overpic}[width=1.0\linewidth]{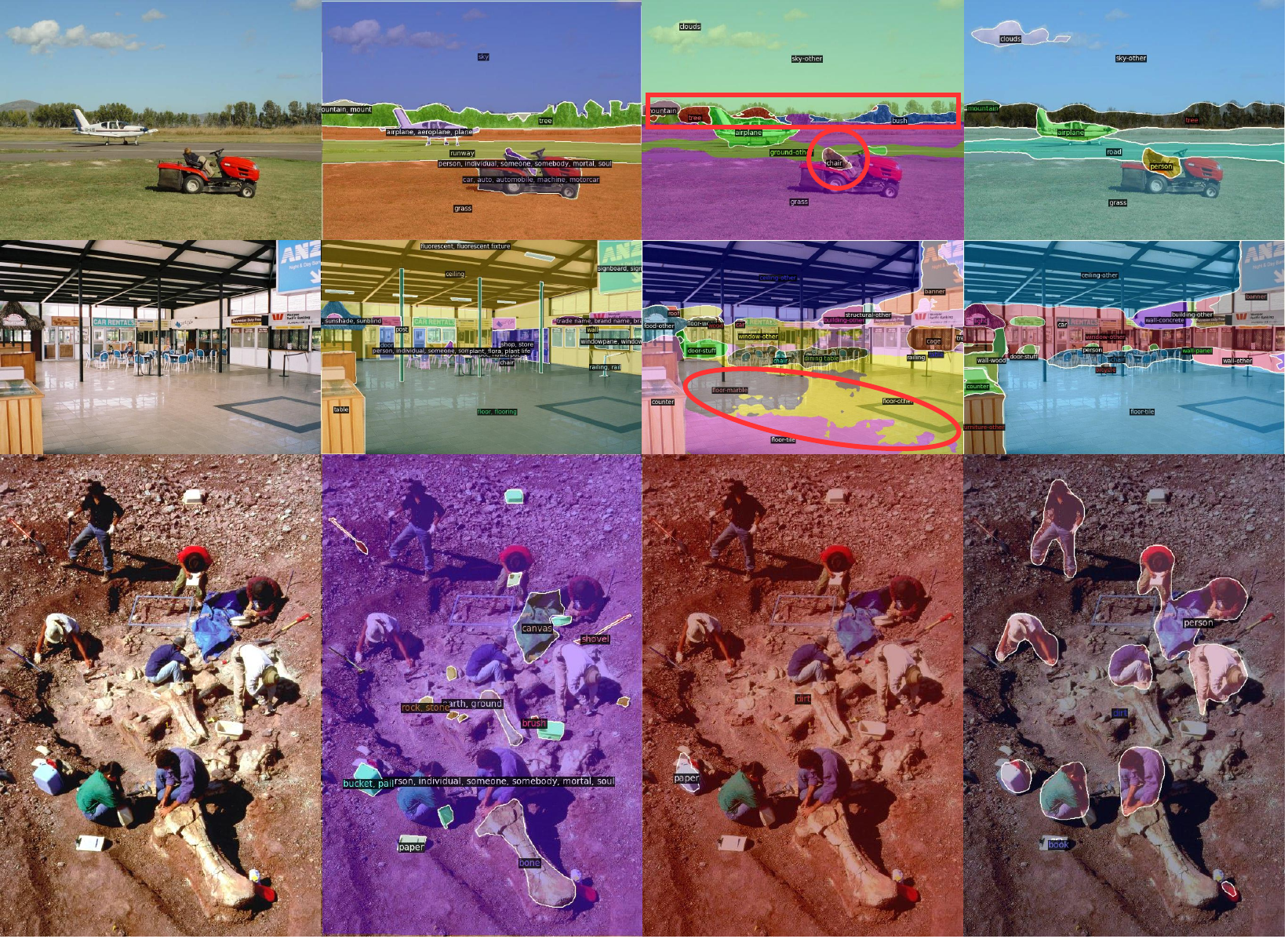}
   \put(10.0,-1.9){{Image}}
   \put(32.5,-1.9){{Ground truth}}
   \put(56.0,-1.9){{HyperCLIP~\cite{hyperclip}}}
   \put(82.5,-1.9){{HyRo (Ours)}}
  \end{overpic}
  \vspace{-11pt}
  \caption{Qualitative comparison between HyperCLIP~\cite{hyperclip} and our method on the \texttt{A-847} setting of ADE20K. Our approach mitigates several semantic misalignment failures observed in HyperCLIP.}
  \vspace{-8pt}
  \label{fig:qualitative}
\end{figure*}

\paragraph{Qualitative Results.}
\Cref{fig:qualitative} presents qualitative comparisons on the ADE20K~\cite{ade20k} dataset, which contains 847 semantic categories (\texttt{A-847}) and serves as the most challenging benchmark in our evaluation due to its highly diverse vocabulary. 

Compared with the state-of-the-art HyperCLIP~\cite{hyperclip}, our proposed HyRo effectively mitigates several common semantic misalignment issues, yielding more cohesive and accurate segmentation masks. In the first row, the baseline fails to distinguish between the person and the chair, incorrectly labeling both as ``chair'' and producing noisy, artifact-heavy predictions along the background tree line. HyRo, however, cleanly disentangles these adjacent objects. In the second row, although HyperCLIP produces a cleaner floor segmentation, it exhibits semantic inconsistency by fragmenting visually uniform regions into multiple contradictory labels, whereas our approach maintains spatial coherence. Finally, in the last row, the baseline succumbs to background dominance, over-predicting the ``dirt'' category across most of the image and completely failing to detect multiple people in the scene. Taken together, these examples highlight HyRo's superior ability to overcome the common failure modes of baseline models, delivering precise and spatially coherent segmentation in highly challenging, open-vocabulary environments.

\vspace{-9pt}

\begin{figure}[t]
    \centering
    \begin{subfigure}[b]{0.325\linewidth}
        \centering
        \includegraphics[width=\linewidth]{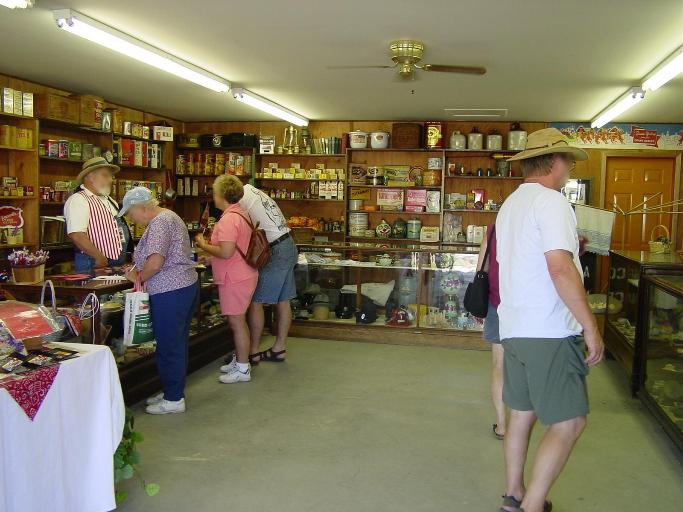}
    \end{subfigure}
    \begin{subfigure}[b]{0.325\linewidth}
        \centering
        \includegraphics[width=\linewidth]{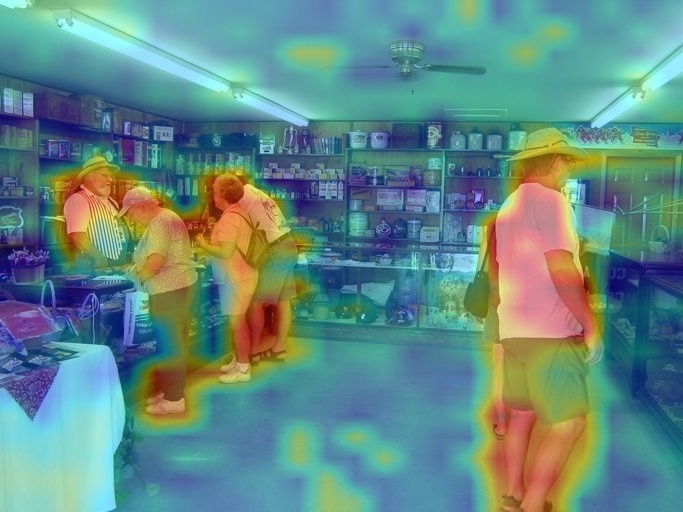}
    \end{subfigure}
    \begin{subfigure}[b]{0.325\linewidth}
        \centering
        \includegraphics[width=\linewidth]{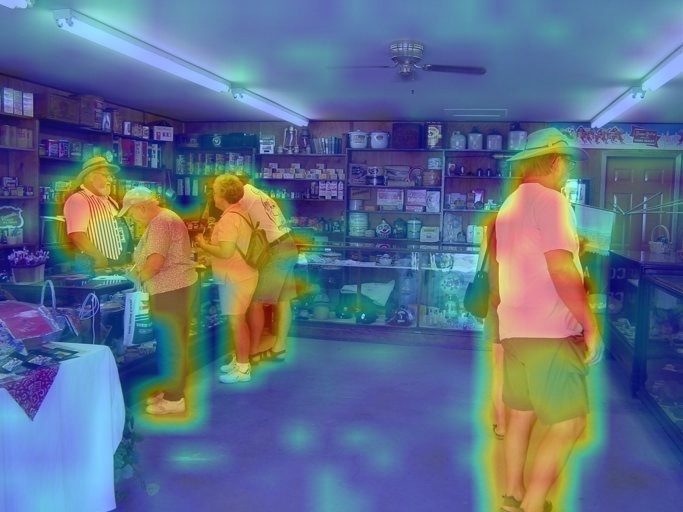}
    \end{subfigure}

    \begin{subfigure}[b]{0.325\linewidth}
        \centering
        \includegraphics[width=\linewidth]{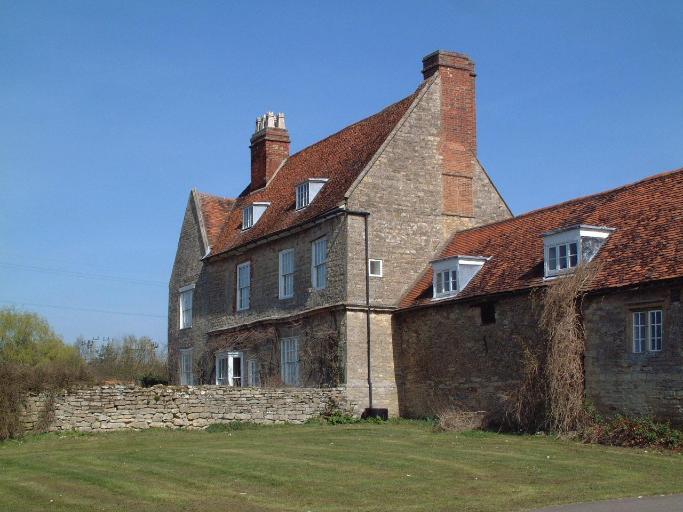}
    \end{subfigure}
    \begin{subfigure}[b]{0.325\linewidth}
        \centering
        \includegraphics[width=\linewidth]{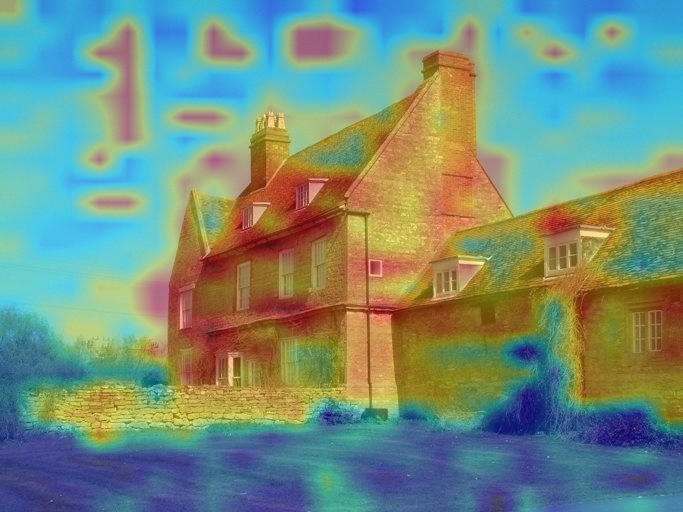}
    \end{subfigure}
    \begin{subfigure}[b]{0.325\linewidth}
        \centering
        \includegraphics[width=\linewidth]{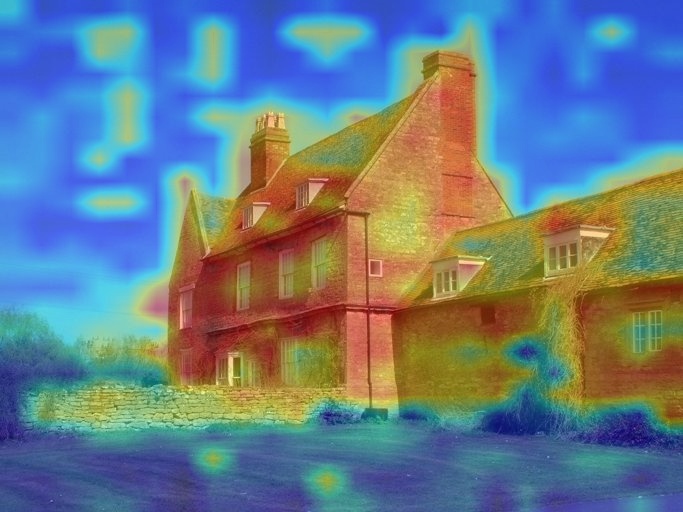}
    \end{subfigure}

    \begin{subfigure}[b]{0.325\linewidth}
        \centering
        \includegraphics[width=\linewidth]{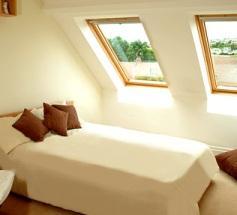}
        \caption{Image}
    \end{subfigure}
    \begin{subfigure}[b]{0.325\linewidth}
        \centering
        \includegraphics[width=\linewidth]{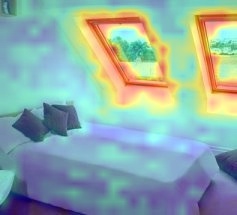}
        \caption{Without HyRo}
    \end{subfigure}
    \begin{subfigure}[b]{0.325\linewidth}
        \centering
        \includegraphics[width=\linewidth]{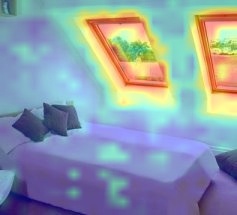}
        \caption{With HyRo (Ours)}
    \end{subfigure}
    
    \vspace{-10pt}
    \caption{Attention map visualization for target classes ``person'', ``building'', and ``window'' (top to bottom). Without HyRo, attention is diffusely spread across semantically irrelevant regions. With HyRo, attention is more concentrated on the target class regions, indicating improved semantic alignment.}
    \label{fig:attn}

    \vspace{-10pt}
\end{figure}

\paragraph{Attention Visualization.}
\Cref{fig:attn} visualizes attention maps of visual embeddings with and without HyRo across three target classes: ``person'', ``building'', and ``window''. Without semantic refinement, the model struggles to ground visual features to the correct class regions, resulting in diffuse attention that bleeds into semantically unrelated background areas. After applying HyRo, the angular refinement in hyperbolic space explicitly improves the semantic correspondence between visual and textual embeddings. This causes attention to concentrate sharply on the precise target regions, effectively suppressing background noise. These visualizations confirm that modeling angular relationships in hyperbolic space is an effective mechanism for enhancing cross-modal semantic alignment.

\subsection{Ablation Study}
\label{subsec:ablation}

\paragraph{Component Importance Analysis.}
We ablate the contributions of radius scaling and rotation in~\cref{tab:ablation_importance}. While independently applying either hierarchical scaling (radius) or semantic refinement (rotation) improves the baseline, their combination in HyRo yields the best performance across nearly all benchmarks. The rotation module provides substantial gains on large-scale sets like \texttt{A-847}, indicating that angular alignment is the primary driver for open-vocabulary generalization. Meanwhile, radius scaling further enhances performance, confirming that positioning embeddings at the correct hierarchical level is essential for fine-grained separability. These results confirm that hierarchical positioning (radius) and angular alignment (rotation) play complementary roles in improving semantic consistency for open-vocabulary segmentation. Specifically, while radius captures abstraction, rotation prevents semantic collapse among closely related categories.
\vspace{-10pt}

\paragraph{Choice of Curvature.} We evaluate the impact of curvature $c$ on performance in~\cref{tab:ablation_curvature}. Results show that a ``gentler" curvature ($c=0.01$) consistently outperforms higher values across most benchmarks. While a higher curvature ($c=1.0$) improves results on specific sets like \texttt{PAS-20$^b$}, it appears to distort the pre-trained Euclidean CLIP space too aggressively for the diverse label space of \texttt{A-847}. Consequently, $c=0.01$ is chosen as the default to preserve CLIP's zero-shot generalization while enabling hierarchical adjustment. We do not explore values smaller than $0.01$, as extremely small curvature would make the space nearly Euclidean and diminish the benefits of hyperbolic geometry.

\paragraph{Size of Diagonal Rotation Blocks.} 
We evaluate the impact of block size $n$ on capturing semantic transformations in~\cref{tab:ablation_diagonal}. Performance scales with $n$, with $n=256$ reaching the optimal balance of representational capacity and generalization across nearly all benchmarks. While smaller datasets like PAS-20 are less sensitive to block size, the significant gains on A-847 suggest that high-capacity rotation is critical for disentangling fine-grained semantic relationships in large-vocabulary settings. Consequently, $n=256$ is selected as the default configuration.

\begin{table}[t]
    \centering
    \resizebox{\linewidth}{!}{
    \begin{tabular}{cc|cccccc}
        \toprule
        Radius & Rotation & \texttt{A-847} & \texttt{PC-459} & \texttt{A-150} & \texttt{PC-59} & \texttt{PAS-20} & \texttt{PAS-20$^b$} \\
        \midrule
        \xmark & \xmark & 11.4 & 17.6 & 29.8 & 56.2 & 94.8 & 75.9 \\
        \cmark & \xmark & 11.9 & 18.2 & \textbf{31.7} & 57.1 & 94.9 & 76.4 \\
        \xmark & \cmark & 11.6 & 18.3 & 30.6 & 56.5 & \textbf{95.4} & \textbf{76.7} \\
        \cmark & \cmark & \textbf{12.0} & \textbf{18.9} & {31.2} & \textbf{57.3} & {95.0} & \textbf{76.7} \\
        \bottomrule
    \end{tabular}
    }
    \vspace{-8pt}
    \caption{Ablation results on the contribution of Radius Scaling and Rotation modules.}
    \label{tab:ablation_importance}
    \vspace{-3pt}
\end{table}

\begin{table}[t]
    \centering
    \resizebox{\linewidth}{!}{
    \begin{tabular}{c|cccccc}
        \toprule
        $c$ & \texttt{A-847} & \texttt{PC-459} & \texttt{A-150} & \texttt{PC-59} & \texttt{PAS-20} & \texttt{PAS-20$^b$} \\
        \midrule
        0.01 & \textbf{12.0} & \underline{18.9} & \textbf{31.2} & \textbf{57.3} & \underline{95.0} & \underline{76.7} \\
        0.05 & \underline{11.4} & \textbf{19.0} & 30.1 & 56.5 & \textbf{95.1} & 76.3 \\
        0.1 & \underline{11.4} & 17.6 & \underline{30.4} & 55.9 & \textbf{95.1} & 75.9 \\
        1.0 & 11.2 & 17.6 & 30.1 & \underline{57.2} & 94.8 & \textbf{77.8} \\
        \bottomrule
    \end{tabular}
    }
    \vspace{-8pt}
    \caption{Ablation results on the choice of curvature $c$.}
    \label{tab:ablation_curvature}
    \vspace{-3pt}
\end{table}

\begin{table}[t]
    \centering
    \resizebox{\linewidth}{!}{
    \begin{tabular}{c|cccccc}
        \toprule
        $n$ & \texttt{A-847} & \texttt{PC-459} & \texttt{A-150} & \texttt{PC-59} & \texttt{PAS-20} & \texttt{PAS-20$^b$} \\
        \midrule
        32 & 11.4 & 17.6 & 29.8 & 56.2 & 94.8 & 75.9 \\
        128 & 11.6 & 18.3 & 30.6 & 56.5 & \textbf{95.4} & \textbf{76.7} \\
        256 & \textbf{12.0} & \textbf{18.9} & \textbf{31.2} & \textbf{57.3} & \underline{95.0} & \textbf{76.7} \\
        \bottomrule
    \end{tabular}
    }
    \vspace{-8pt}
    \caption{Ablation results on the size of diagonal rotation blocks $n$.}
    \label{tab:ablation_diagonal}
    \vspace{-10pt}
\end{table}

\section{Conclusion}
\label{sec:conclusion}
\vspace{-2pt}
In this work, we introduce HyRo, a fine-tuning strategy for open-vocabulary semantic segmentation in hyperbolic space that explicitly decouples hierarchical alignment and semantic refinement. By operating in the Poincar\'e ball model, HyRo first aligns embeddings to appropriate hierarchical levels via radius adjustment and then refines semantic relationships through angular optimization using orthogonal transformations that preserve the radius. This strategy enables more precise alignment between vision and language representations without affecting generalization to unseen categories. Experiments on multiple benchmarks demonstrate that HyRo consistently improves performance over strong baselines. This work highlights the importance of jointly modeling hierarchy and semantics in non-Euclidean spaces and opens new directions for geometric representation learning in dense vision--language tasks.
\vspace{-10pt}

\paragraph{Future Work.} While HyRo demonstrates strong open-vocabulary segmentation performance on static images, extending this hyperbolic geometric approach to more complex open-vocabulary video segmentation settings, such as MOSE~\cite{MOSE}, MOSEv2~\cite{MOSEv2}, MeViS~\cite{MeViS}, and MeViSv2~\cite{MeViSv2}, is a promising direction for future work.

{
    \small
    \bibliographystyle{ieeenat_fullname}
    \bibliography{main}
}


\end{document}